\documentclass[conference]{IEEEtran}
\IEEEoverridecommandlockouts

\usepackage{cite}
\usepackage{amsmath,amssymb,amsfonts}
\usepackage{algorithmic}
\usepackage{graphicx}
\usepackage{textcomp}
\usepackage{xcolor}
\usepackage{bm}

\def\BibTeX{{\rm B\kern-.05em{\sc i\kern-.025em b}\kern-.08em
    T\kern-.1667em\lower.7ex\hbox{E}\kern-.125emX}}

\begin{document}

\title{SoftNash: Entropy-Regularized Nash Games for Non-Fighting Virtual Fixtures}

\author{\IEEEauthorblockN{1\textsuperscript{st} Tai Inui}
\IEEEauthorblockA{\textit{Waseda University}\\
Tokyo, Japan\\
0009-0000-2387-2790}
\and
\IEEEauthorblockN{2\textsuperscript{nd} Jee-Hwan Ryu}
\IEEEauthorblockA{\textit{KAIST}\\
Daejeon, South Korea\\
0000-0002-6497-7115}
}

\maketitle

\begin{abstract}
Virtual fixtures (VFs) improve precision in teleoperation but often ``fight'' the user, inflating mental workload and eroding the sense of agency. We propose \emph{Soft-Nash Virtual Fixtures}, a game-theoretic shared-control policy that softens the classic two-player linear-quadratic (LQ) Nash solution by inflating the fixture's effort weight with a single, interpretable scalar parameter~$\tau$. This yields a continuous dial on controller assertiveness: $\tau=0$ recovers a hard, performance-focused Nash / virtual fixture controller, while larger $\tau$ reduce gains and pushback, yet preserve the equilibrium structure and continuity of closed-loop stability. We derive Soft-Nash from both a KL-regularized trust-region and a maximum-entropy viewpoint, obtaining a closed-form robot best response that shrinks authority and aligns the fixture with the operator's input as $\tau$ grows. We implement Soft-Nash on a 6-DoF haptic device in 3D tracking task (n=12). Moderate softness ($\tau\approx 1$--$3$, especially $\tau=2$) maintains tracking error statistically indistinguishable from a tuned classic VF while sharply reducing controller-user conflict, lowering NASA-TLX workload, and increasing Sense of Agency (SoAS). A composite BalancedScore that combines normalized accuracy and non-fighting behavior peaks near $\tau=2$--3. These results show that a one-parameter Soft-Nash policy can preserve accuracy while improving comfort and perceived agency, providing a practical and interpretable pathway to personalized shared control in haptics and teleoperation.
\end{abstract}


\section{Introduction}

Teleoperation enables people to act in remote or hazardous environments by commanding a robot through an interface. \emph{Virtual fixtures} (VFs) are haptic or kinematic constraints that overlay task-specific guidance onto the operator's input, such as snapping motion along a desired path or preventing entry into forbidden regions~\cite{rosenberg93virtual,rosenberg93report}. In surgical robotics, manufacturing, and maintenance tasks, VFs and related active constraints can substantially reduce error and improve safety by stabilizing motion and enforcing geometric constraints.

A persistent challenge, however, is that conventional VFs often behave like rigid ``hard'' constraints: high-gain impedance fields that keep the tool on track at all costs. When the operator deliberately deviates for exploration, strategy, or recovery, the fixture can apply strong opposing forces, causing the human and robot to \emph{fight} each other. This controller--user conflict increases physical and mental workload, may degrade performance when intent is misinterpreted, and can erode trust and the sense that ``I am the one doing this''~\cite{du2014flexible,ni2017haptic}. Human-centered design guidelines emphasize that automation should support, not override, the human; assistance that feels over-assertive can harm acceptance and skill acquisition~\cite{parasuraman00model}.

To mitigate this, prior work has explored \emph{soft} or \emph{compliant} VFs that reduce stiffness, modulate gains, or adapt forces based on performance~\cite{du2014flexible,ni2017haptic}. In parallel, formal models of human--robot shared control have used optimal control and game theory, treating the human and robot as players in a differential game that each minimize a quadratic cost~\cite{jarrasse2012framework,music2020haptic,zou2020framework}. Such models interpret classic master--slave and teacher--student assistance schemes as different allocations of cost and authority, and have been extended to shared autonomy in high-DOF teleoperation where robot assistance is blended with user input~\cite{dragan2013policyblending,javdani2018shared}.

Despite these advances, tuning the trade-off between robot assistance and human freedom remains an open design problem. In practice, designers often adjust virtual stiffness, damping, or blending gains by trial and error. Linear blending of human and robot commands yields a single mixing parameter, but it is usually heuristic and does not arise from a principled optimality criterion; changing it can subtly affect stability and does not expose how, or where, the robot should yield. Likewise, in game-theoretic formulations the robot's aggressiveness is typically encoded implicitly in cost matrices; modifying them can disrupt equilibrium structure and lacks a single, interpretable parameter for ``how hard the fixture pushes.''

We introduce Soft-Nash Virtual Fixtures to address this gap. We model the interaction as a two-player linear-quadratic (LQ) game between a human and a robot assistant, and add an entropy/KL regularization term to the robot's cost that encourages consistency with a prior policy anchored to the human. In the linear--Gaussian setting, this regularization yields a closed-form robot best response whose feedback gains depend on a single scalar softness parameter $\tau \ge 0$. At $\tau = 0$ we recover a hard Nash / classic VF solution; as $\tau \to \infty$, the robot's influence vanishes and the system approaches pass-through of the human input. Intermediate values of $\tau$ interpolate smoothly between these extremes while preserving equilibrium existence and stability.

SoftNash is intended as a non-fighting shared-control primitive for haptic teleoperation. Rather than blending a pre-defined robot command with the human's input, SoftNash reshapes the underlying game: the same quadratic objectives are optimized under an entropy-regularized Nash formulation, and the scalar parameter $\tau$ directly controls how strongly the VF enforces its geometric preferences. This yields an interpretable knob for trading off performance-maximizing guidance against agency-preserving compliance, complementary to task-specific design of potential fields or safety constraints. While our experiments focus on a 3D tracking task with a grounded haptic device, the formulation applies more broadly to VF-style constraints in telemanipulation.

In summary, this paper makes the following contributions:
\begin{itemize}
    \item We formulate a two-player LQ game between a human and a robot assistant and introduce an entropy/KL regularization in the robot's cost that anchors its policy to a human-referenced prior. In the linear--Gaussian setting, we derive a closed-form best response and show that the regularization strength $\tau \ge 0$ acts as a softness knob: $\tau=0$ recovers a hard Nash / classic VF solution; $\tau \to \infty$ approaches pass-through of the human input; intermediate $\tau$ values smoothly trade off assistance and compliance.
    \item We analyze how SoftNash differs from simple gain scaling or linear blending, and show that it preserves the equilibrium structure and stability guarantees of LQ games while providing a single interpretable parameter for ``how hard the fixture pushes.''
    \item We implement SoftNash on a 6-DoF haptic device (Force Dimension Omega.6) for a 3D tracking task, and define metrics for tracking accuracy, interaction energy, a Non-Fighting Index (NFI) that normalizes conflict by delivered assistance, and a BalancedScore combining accuracy and non-fighting behavior.
    \item In an $n=12$ within-subjects study across eight assistance modes, we show that moderate softness (around $\tau \approx 2$) maintains accuracy comparable to a tuned classic VF while substantially reducing controller--user conflict and workload, and improving perceived agency.
\end{itemize}

Together, these results demonstrate that a single interpretable parameter can continuously tune shared control between performance-maximizing guidance and agency-preserving compliance, providing a practical design handle for non-fighting virtual fixtures.

\section{Background and Related Work}

\subsection{Virtual Fixtures and Soft Constraints}

Virtual fixtures were introduced as perceptual tools for telerobotic manipulation, showing that virtual guides and forbidden-region fields can significantly reduce task time and error in peg-in-hole teleoperation~\cite{rosenberg93virtual,rosenberg93report}. Subsequent work extended VFs to surgical and industrial settings, including flexible or adaptive fixtures whose impedance depends on task context~\cite{du2014flexible,ni2017haptic}. These approaches highlight the trade-off between tight guidance for accuracy and compliance for comfort and exploration.

Soft or compliant fixtures typically replace hard constraints with virtual springs and dampers or modulate guidance gains based on error, velocity, or inferred intent. While effective, such designs often rely on heuristic gain scheduling; explicitly connecting these heuristics to optimality or robustness guarantees is non-trivial.

\subsection{Game-Theoretic Shared Control}

Game-theoretic models treat human--robot interaction as a dynamic game in which each agent optimizes its own objective subject to shared dynamics~\cite{jarrasse2012framework,zou2020framework}. Prior work has classified interaction patterns such as master--slave, teacher--student, and collaborative schemes by how control costs and task penalties are assigned to human and robot~\cite{jarrasse2012framework}. More recent work applies $N$-player differential LQ games to human--robot(-human) physical interaction, using estimated human objectives to adapt robot control online~\cite{zou2020framework,music2020haptic}.

In teleoperation and shared autonomy, policy-blending formulations blend user and robot policies based on predicted goals and desired assistance levels~\cite{dragan2013policyblending}, and shared autonomy has been framed as hindsight-optimized assistance under uncertain user goals~\cite{javdani2018shared}. These approaches emphasize prediction and arbitration but do not provide a single global parameter for softening a Nash solution while preserving its structure.

\subsection{Subjective Measures: Agency and Workload}

To evaluate perceived control and workload, we draw from psychological and human factors measures. The Sense of Agency Scale (SoAS) measures individuals' beliefs about their general control over body, mind, and environment~\cite{tapal2017soas}. NASA's Task Load Index (NASA-TLX) is a widely used multidimensional rating scale for subjective workload, spanning mental and physical demand, temporal demand, performance, effort, and frustration~\cite{hart1988nasa}. We adapt SoAS items to assess perceived agency in physical teleoperation, and employ unweighted NASA-TLX to summarize workload across assistance modes.

\section{Soft-Nash Virtual Fixtures}

\subsection{Problem Setting}

We consider a discrete-time teleoperation setting with linear task dynamics and an additive robot control channel. Let $x_k\in\mathbb{R}^n$ denote the task-relevant state at time step $k$, and let $u_{r,k}\in\mathbb{R}^m$ be the robot-assistive control applied through a known input matrix $B$:
\begin{equation}
    x_{k+1} = A x_k + B u_{r,k}, \qquad k = 0,1,\dots
    \label{eq:dynamics}
\end{equation}
The human provides a command $u_{h,k}\in\mathbb{R}^m$, for example a desired Cartesian velocity from a haptic device. We define the tracking error against a reference trajectory $r_k$ as
\begin{equation}
    e_k = C x_k - r_k,
\end{equation}
with $C$ selecting task outputs (e.g., tool position). We assume standard quadratic penalties on state error and control effort with $Q_r\succeq 0$, $Q_h\succeq 0$, and $R_r\succ 0$, $R_h\succ 0$.

\subsection{Stage Costs and Two-Player Game}

We model the robot and human as LQ players in a stage game coupled through $x_k$ and $e_k$. The robot's stage loss encodes tracking performance, control energy, and a soft attraction of its action toward a human-anchored reference:
\begin{equation}
\begin{split}
    \ell_r(x_k,u_{r,k};u_{h,k})
    = {} & e_k^\top Q_r e_k 
          + u_{r,k}^\top R_r u_{r,k} \\
        & {}+ \tau (u_{r,k} - \alpha u_{h,k})^\top
              S (u_{r,k} - \alpha u_{h,k})
\end{split}
\label{eq:robot-cost}
\end{equation}

where $S\succ 0$ defines a metric in action space, $\alpha\in\mathbb{R}$ sets a nominal blending factor with the human command, and $\tau\ge 0$ is the \emph{softness} parameter.

For analysis we also define a human quadratic loss
\begin{equation}
    \ell_h(x_k,u_{h,k};u_{r,k}) 
    = e_k^\top Q_h e_k + u_{h,k}^\top R_h u_{h,k},
    \label{eq:human-cost}
\end{equation}
though in experiments the human may behave in a reactive or exogenous way rather than optimally minimizing~\eqref{eq:human-cost}.

Together with the dynamics~\eqref{eq:dynamics}, these losses define a two-player LQ dynamic game. Under standard stabilizability/detectability assumptions, the resulting Nash equilibrium can be characterized by Riccati equations for each player.

In Soft-Nash VFs we focus on the robot \emph{best response} given $x_k$ and $u_{h,k}$, and treat the human command as an exogenous input. The key question is how the regularization parameter $\tau$ shapes this best response.

\subsection{Regularization via Trust Region and Maximum Entropy}

The additional term in~\eqref{eq:robot-cost} arises from two equivalent viewpoints that regularize the robot's policy relative to a human-anchored prior.

\subsubsection{KL-Trust-Region View (Deterministic Limit)}

Consider a stochastic robot policy $\pi(u_r\mid x,u_h)$ and a prior Gaussian policy centered on a scaled human command,
\begin{equation}
    \pi_0(u_r \mid x,u_h) = \mathcal{N}(\alpha u_h, \Sigma_0).
\end{equation}
At a given state $(x,u_h)$ we solve a regularized stage problem
\begin{equation}
    \min_{\pi} \; \mathbb{E}_{\pi}[\ell_r(x,u_r;u_h)]
    + \beta \,\mathrm{KL}\bigl(\pi \,\|\, \pi_0\bigr),
    \label{eq:kl-problem}
\end{equation}
with weight $\beta>0$. In the deterministic limit where $\pi$ collapses to a Dirac at some $u_r^\star$, the KL divergence becomes a quadratic penalty on deviations from $\alpha u_h$:
\begin{equation}
\begin{split}
    \mathrm{KL}\bigl(\delta(\cdot-u_r^\star)&\,\|\,\mathcal{N}(\alpha u_h,\Sigma_0)\bigr)
    = \\
    &\tfrac{1}{2}(u_r^\star-\alpha u_h)^\top
      \Sigma_0^{-1}(u_r^\star-\alpha u_h)
      + \mathrm{const}.
\end{split}
\end{equation}

Substituting into~\eqref{eq:kl-problem}, the deterministic best response solves
\begin{equation}
    \min_{u_r} \; e^\top Q_r e 
    + u_r^\top R_r u_r
    + \tfrac{\beta}{2} (u_r - \alpha u_h)^\top \Sigma_0^{-1} (u_r - \alpha u_h).
\end{equation}
Identifying
\begin{equation}
    S = \tfrac{1}{2}\Sigma_0^{-1}, 
    \qquad
    \tau = \beta,
\end{equation}
we recover the soft attraction term in~\eqref{eq:robot-cost}. Larger $\tau$ corresponds to a tighter trust region that keeps the robot action closer to the human-anchored prior; smaller $\tau$ recovers the unregularized Nash solution.

\subsubsection{Maximum-Entropy View}

Alternatively, consider a maximum-entropy control problem
\begin{equation}
    \min_{\pi} \; \mathbb{E}_{\pi}[\ell_r(x,u_r;u_h)] 
    + \beta\,\mathrm{KL}(\pi\|\pi_0) 
    - \lambda H(\pi),
\end{equation}
where $H(\pi)$ is the differential entropy of $\pi$. For quadratic $\ell_r$ and Gaussian $\pi_0$, the optimizer is Gaussian $\pi^\star=\mathcal{N}(\mu^\star,\Sigma^\star)$ with
\begin{align}
    \mu^\star 
    &= \arg\min_{u_r} \; e^\top Q_r e + u_r^\top R_r u_r \notag\\
    &\quad + \beta (u_r-\alpha u_h)^\top \Sigma_0^{-1} (u_r-\alpha u_h), \\
    \Sigma^\star 
    &= \bigl(2 R_r + 2\beta \Sigma_0^{-1}\bigr)^{-1}\cdot(2\lambda),
\end{align}

so the mean again solves a quadratic problem with attraction toward $\alpha u_h$, and the entropy term only inflates covariance. Taking $\lambda\to 0$ recovers the deterministic trust-region solution above. In both viewpoints, $\tau=\beta$ controls how strongly the robot matches the prior \emph{mean} while preserving LQ structure.

\subsection{Closed-Form Robot Best Response}

Including the value function of the dynamic game, the robot's total cost-to-go is approximated as $V_r(x)=x^\top P_r x$ with $P_r\succeq 0$ the solution of a discrete-time Riccati equation. Substituting one-step lookahead with $x_{k+1}$ from~\eqref{eq:dynamics} and $V_r(x_{k+1})$ into the Bellman equation yields a quadratic minimization in $u_{r,k}$.

Collecting terms, the Hessian and linear term in $u_{r,k}$ are
\begin{align}
    H(\tau) &= R_r + \tau S + B^\top P_r B, \label{eq:Htau}\\
    b(x_k,u_{h,k},\tau) &= B^\top P_r A x_k - \tau \alpha S u_{h,k}.
    \label{eq:btau}
\end{align}
Assuming $H(\tau)\succ 0$ (guaranteed for $\tau\ge 0$ with $R_r\succ 0$, $S\succ 0$, and $P_r\succeq 0$), the robot best response is
\begin{equation}
    u_{r,k}^\star(x_k,u_{h,k};\tau) 
    = -H(\tau)^{-1} b(x_k,u_{h,k},\tau).
    \label{eq:best-response}
\end{equation}

\subsubsection{Comparative Statics in $\tau$}

From~\eqref{eq:Htau}--\eqref{eq:best-response} we can interpret the effect of $\tau$:

\paragraph*{Shrinking authority}
As $\tau$ increases, $H(\tau)$ grows in the direction of $S$, and its inverse shrinks. This reduces the magnitude of the state feedback gain
\begin{equation}
    K_r(\tau) = H(\tau)^{-1} B^\top P_r A,
\end{equation}
softening the fixture's corrective action on $x_k$.

\paragraph*{Human alignment}
The affine term
\begin{equation}
    \alpha(\tau) = H(\tau)^{-1} (\tau \alpha S)
\end{equation}
scales the contribution of $u_{h,k}$ in~\eqref{eq:best-response}. For small $\tau$, $\alpha(\tau)$ is near zero and the controller largely ignores the human input beyond dynamics; as $\tau$ grows, $\alpha(\tau)$ approaches $\alpha I$, making the controller increasingly consistent with the human command.

\paragraph*{Limiting cases}
\begin{itemize}
    \item \textbf{Hard-Nash / classic VF ($\tau=0$)}: The best response reduces to the standard LQ Nash / VF controller that optimizes tracking and effort without explicit regularization toward $u_h$:
    \begin{equation}
        u_{r,k}^\star = -\bigl(R_r + B^\top P_r B\bigr)^{-1} B^\top P_r A x_k.
    \end{equation}
    \item \textbf{Very soft / pass-through ($\tau\to\infty$)}: The attraction term dominates and $H(\tau)^{-1}\tau S\to I$, so $u_{r,k}^\star\to\alpha u_{h,k}$ modulo small corrections from $Q_r$ and $P_r$. The robot behaves as a scaled pass-through of the human command.
\end{itemize}

\subsection{Closed-Loop Dynamics and Stability}

If we model the human as a state-feedback controller $u_{h,k}=K_h x_k$ (or as bounded exogenous input), substituting~\eqref{eq:best-response} into~\eqref{eq:dynamics} yields
\begin{equation}
    x_{k+1} = \bigl( A - B K_r(\tau) \bigr) x_k + B \alpha(\tau) K_h x_k.
\end{equation}
For small to moderate $\tau$, $K_r(\tau)$ retains enough authority to stabilize $(A,B)$ in combination with $K_h$, and standard LQ game theory ensures existence of a stabilizing Nash equilibrium. As $\tau$ becomes very large, $K_r(\tau)\to 0$ and closed-loop stability increasingly depends on the human; the system gradually reduces to human-only control with pass-through assistance.

Soft-Nash VFs thus provide a continuous, interpretable interpolation between a hard, performance-focused fixture and a fully permissive pass-through, while preserving the equilibrium structure of the underlying game.

\section{Implementation on a 6-DoF Haptic Device}

We implement Soft-Nash VFs on a Force Dimension Omega.6, a grounded 6-DoF haptic interface with a pen-style end-effector. The user controls a virtual stylus position $\bm{p}(t)\in\mathbb{R}^3$ in a bounded workspace, and the controller renders assistive forces $\bm{F}_a(t)$ at $\sim$1~kHz.

\subsection{Control Architecture}

We discretize the stylus dynamics at period $T$ and treat the end-effector position as the state:
\begin{equation}
    x_k = 
    \begin{bmatrix}
        \bm{p}_k \\ \dot{\bm{p}}_k
    \end{bmatrix}, 
    \qquad
    u_{r,k} = \bm{F}_a(kT).
\end{equation}
A second-order mass--damper model yields $A$ and $B$ for~\eqref{eq:dynamics}. The human command $u_{h,k}$ is taken as the unconstrained output of a baseline impedance controller or as the raw device velocity scaled by a gain. The classic VF mode corresponds to a high-gain PD controller on the tracking error $e_k$.

Soft-Nash assistance replaces the classic VF force with~\eqref{eq:best-response}, evaluated at each servo step with precomputed $P_r$ and $H(\tau)$ for chosen $\tau$ values. In all modes we clip forces to a safe bound and fade forces smoothly at trial boundaries.

\subsection{Target Trajectory}

The target position $\bm{q}(t)\in\mathbb{R}^3$ follows a smooth pseudo-random trajectory generated by a bounded random walk with low-pass filtering and workspace saturation. Each 60~s trial uses a seed-specific but deterministic path, ensuring reproducibility while avoiding exact repetition across participants.

\begin{figure}[t]
    \centering
    \includegraphics[width=0.9\linewidth]{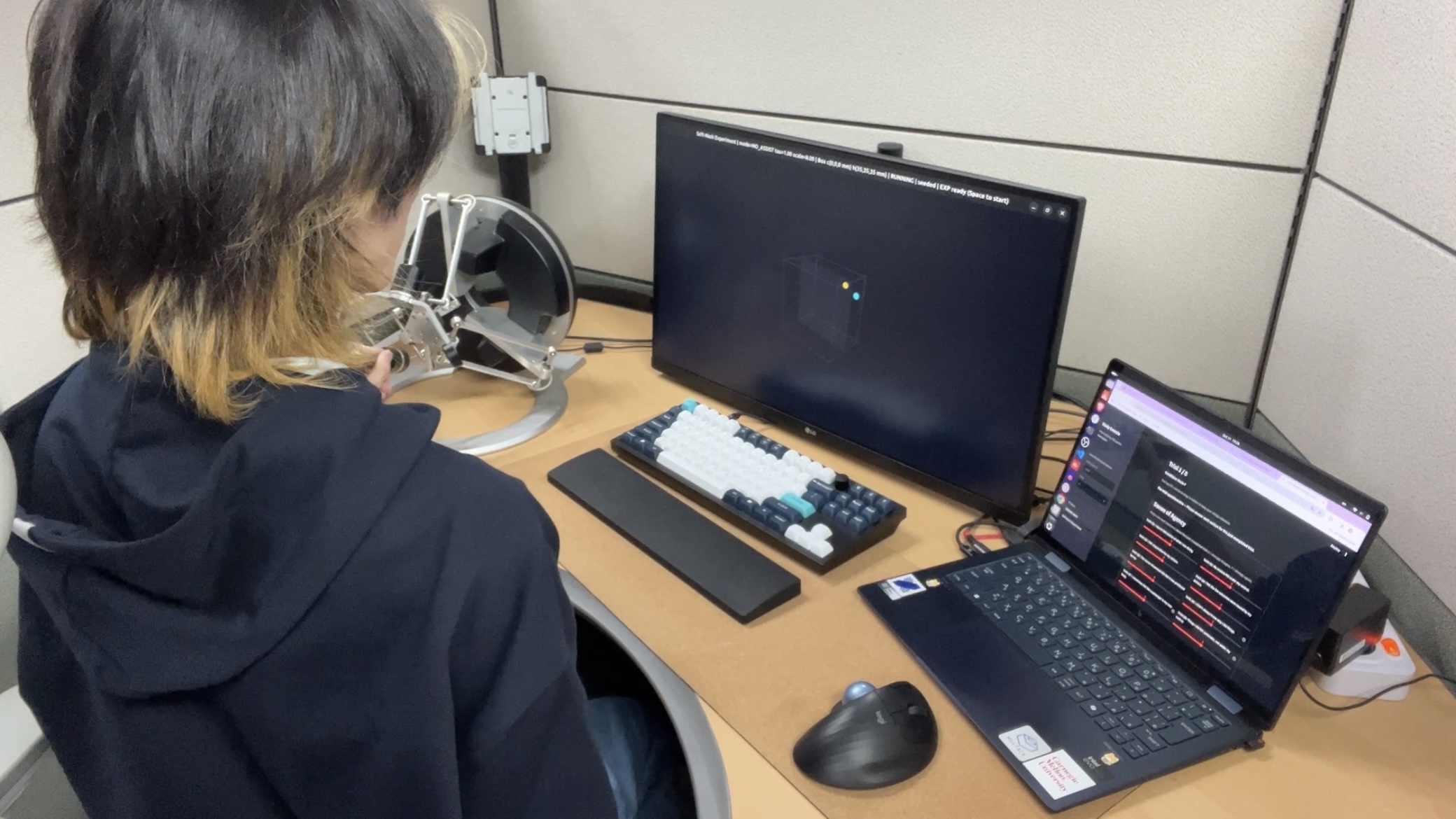}
    \caption{Experiment Setup. Each participant performed a 60 second 3D tracking task with a haptic device followed by a questionnaire.}
    \label{fig:subjective-results}
\end{figure}

\section{User Study Design}

\subsection{Participants and Design}

Twelve right-handed volunteers (n=12) completed a single-session study. The design was within-subjects with eight assistance modes:
\begin{equation}
    M = \{\text{CLASSIC}, \text{NASH}_{\tau\in\{0,1,2,3,5,8\}}, \text{NONE}\}.
\end{equation}
Here, \texttt{CLASSIC} denotes a tuned conventional VF controller, \texttt{NASH\_0} is the non-regularized hard-Nash controller ($\tau=0$), \texttt{NASH\_$\tau$} for $\tau\in\{1,2,3,5,8\}$ are Soft-Nash modes, and \texttt{NONE} is pure teleoperation with no assistive forces.

Each participant experienced all modes in a counterbalanced order drawn from a pre-generated schedule to mitigate sequence and learning effects.

\subsection{Task and Procedure}

In each trial, participants used the Omega.6 stylus to track the moving 3D target point for 60~s. The cursor (stylus) and target were displayed in real time on a monitor. Instructions emphasized both accuracy and natural movement: participants were asked to keep the cursor centered on the target as closely as possible without over-focusing on micro-corrections.

Participants completed a short familiarization block before data collection. For each mode, they then performed one 60~s tracking trial followed by questionnaires. Short breaks were provided between trials to reduce fatigue.

After each trial, participants completed:
\begin{itemize}
    \item A Sense of Agency Scale (SoAS) questionnaire adapted for teleoperation (10 items on a 1--7 Likert scale)~\cite{tapal2017soas}.
    \item The NASA-TLX workload questionnaire (6 subscales, 0--100 visual analog scales)~\cite{hart1988nasa}, using the unweighted average as a summary score.
\end{itemize}

\subsection{Objective Metrics}

We define four objective metrics to characterize accuracy and interaction behavior.

\subsubsection{RMS Tracking Error}

We quantify tracking accuracy by the root mean squared (RMS) distance between stylus and target over a trial of duration $T$:
\begin{equation}
    \mathrm{RMS} = 
    \Biggl(
        \frac{1}{T} \int_0^T \| \bm{e}(t) \|^2 \, dt
    \Biggr)^{1/2},
    \qquad
    \bm{e}(t) = \bm{p}(t) - \bm{q}(t).
\end{equation}
Lower RMS indicates closer tracking.

\subsubsection{Conflict / Interaction Energy}

To capture how much the controller opposes the user's motion, we compute the time integral of the positive part of the negative power exchange between the assistive force and the end-effector velocity:
\begin{equation}
    E_{\mathrm{conflict}} 
    = \int_0^T \max\bigl(0,\,-\bm{F}_a(t)\cdot\bm{v}(t)\bigr)\,dt,
\end{equation}
where $\bm{v}(t)$ is the stylus velocity. This accumulates energy only when controller and user do work against each other. Lower values indicate more cooperative, non-fighting assistance.

\subsubsection{Non-Fighting Index (NFI)}

Raw conflict energy can be small either because the controller is gentle or because it barely assists. To normalize conflict by delivered help, we define
\begin{equation}
    E_{\mathrm{assist}} = \int_0^T \|\bm{F}_a(t)\|\,dt, \qquad
    \mathrm{NFI} = \frac{E_{\mathrm{conflict}}}{E_{\mathrm{assist}}}.
\end{equation}
Lower NFI indicates more cooperative assistance per unit of delivered force: comparable assistance with less conflict (desirable), or conversely, disproportionate conflict for the assistance budget (undesirable).

\subsubsection{BalancedScore}

To provide a single composite measure without tuning weights in physical units, we compute BalancedScore by min--max normalizing RMS error and conflict energy across all trials:
\begin{align}
    p &= \frac{\mathrm{RMS} - \min(\mathrm{RMS})}
              {\max(\mathrm{RMS}) - \min(\mathrm{RMS})}, \\
    f &= \frac{E_{\mathrm{conflict}} - \min(E_{\mathrm{conflict}})}
              {\max(E_{\mathrm{conflict}}) - \min(E_{\mathrm{conflict}})},
\end{align}
and then averaging the complements:
\begin{equation}
    \mathrm{BalancedScore}
    = \tfrac{1}{2}(1-p) + \tfrac{1}{2}(1-f).
\end{equation}
BalancedScore treats accuracy and non-fighting as co-primary objectives; higher is better. The normalization prevents either dimension from dominating due to scale.

\section{Results}

We analyzed objective metrics and subjective scores using linear mixed-effects models (participant as a random intercept) or repeated-measures ANOVA, with assistance mode as a fixed effect. Unless noted, post-hoc tests used Holm-adjusted $p$-values to control the family-wise error rate across planned contrasts.

\subsection{Tracking Accuracy}

\begin{figure}[t]
    \centering
    \includegraphics[width=\linewidth]{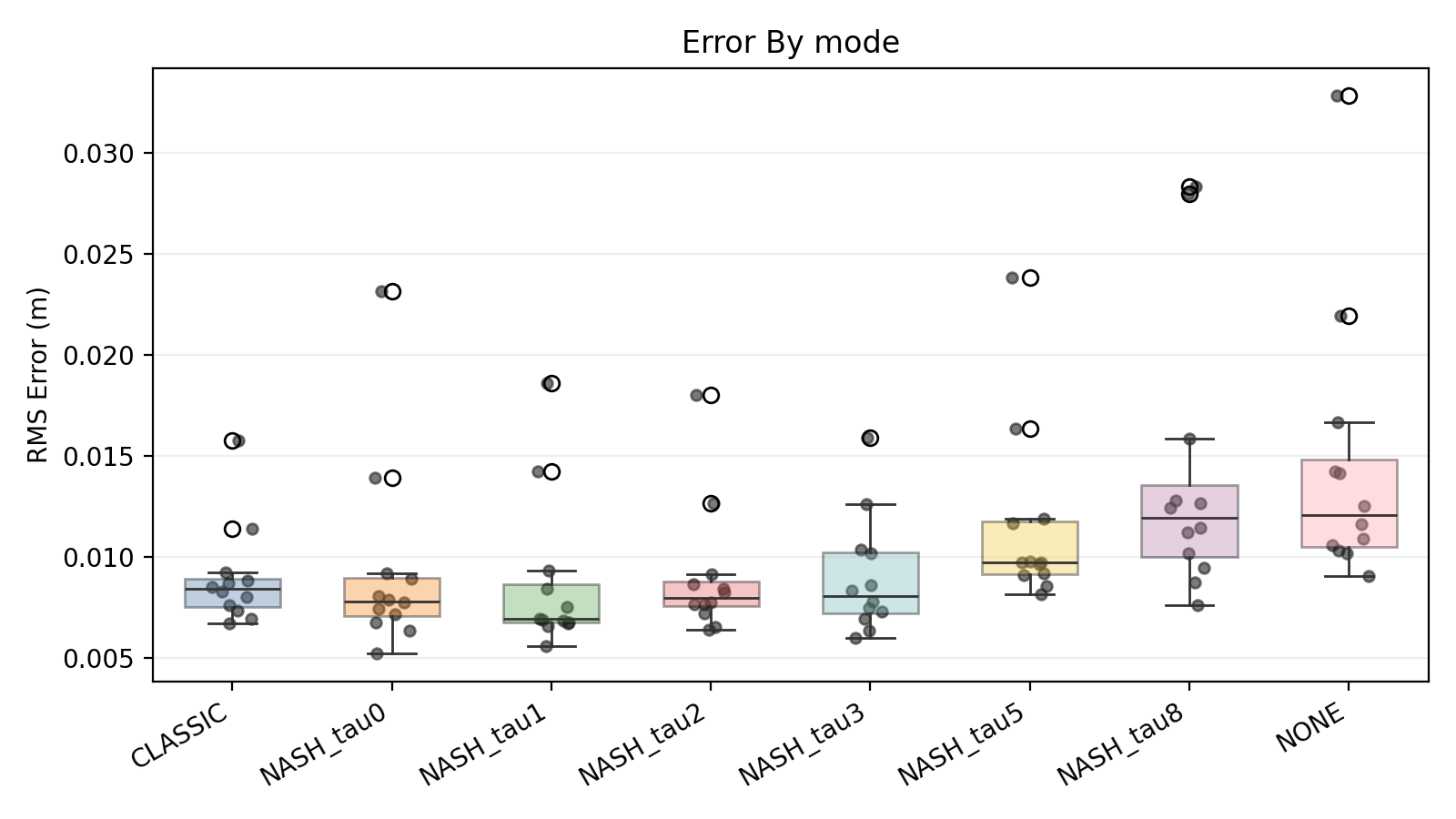}
    \caption{RMS tracking error by assistance mode (participant means $\pm$95\% CI). Classic VF and SoftNash with $\tau \le 3$ achieve similar error, while very soft controllers ($\tau \ge 5$) drift toward the no-assistance baseline.}
    \label{fig:rms-error}
\end{figure}

We fit a one-way mixed-effects model for RMS tracking error (96 observations from 12 participants; model converged). Relative to \texttt{NONE}, classic VF reduced error by $5.63$~mm (difference in means; $z=-3.06$, $p=0.0022$, Holm-adjusted $p=0.0287$). SoftNash modes with $\tau \in \{0,1,2,3\}$ all had mean RMS errors around $9$~mm and were statistically indistinguishable from classic VF (all $|z|\le 0.20$, all Holm-adjusted $p \ge 0.84$). 

For very soft controllers, error increased toward the no-assistance level: $\tau=5$ (mean $\approx 11$~mm) was not significantly different from classic VF ($z=1.37$, $p=0.17$, Holm $p=1.0$), while $\tau=8$ (mean $\approx 14$~mm) was significantly worse than classic VF ($z=2.78$, $p=0.0055$, Holm $p=0.0439$) and close to \texttt{NONE} (mean $\approx 15$~mm; $z=-0.28$, $p=0.78$). Overall, moderate SoftNash gains ($\tau \le 3$) preserved classic VF–level accuracy, whereas extremely large $\tau$ degraded performance.

\subsection{Conflict Energy and NFI}

Conflict energy showed a robust main effect of mode in a one-way repeated-measures ANOVA ($F(7,77)=13.38$, $p<0.001$). Classic VF yielded a mean conflict energy of $0.053$~J (95\% CI $[0.027,\,0.079]$), whereas hard-Nash ($\tau=0$) reduced this to $0.040$~J, approximately a 25\% reduction. SoftNash with $\tau\in\{1,2,3\}$ further reduced conflict energy to $0.020$, $0.012$, and $0.017$~J, respectively (i.e., $\sim 62$–$77$\% below classic VF). For very soft fixtures ($\tau\ge 5$), conflict energy approached the floor set by \texttt{NONE} (mean $\le 0.014$~J vs.\ $0$~J for \texttt{NONE}), indicating that the controller engaged infrequently and seldom pushed against the user.

To factor out the amount of force actually delivered, we modeled the Non-Fighting Index (NFI) with a mixed-effects model. Classic VF and hard-Nash produced mean NFI $\approx 0.002$. Compared to classic VF, SoftNash at $\tau=1$, $2$, and $3$ significantly reduced NFI (all $z \le -2.92$, nominal $p \le 0.0036$, Holm-adjusted $p<0.05$), halving NFI to $\sim 0.001$ on average. SoftNash at $\tau=5$ showed a nominal reduction in NFI ($z=-2.16$, $p=0.031$) that did not survive Holm correction (adjusted $p=0.185$), and $\tau=8$ trended higher than classic VF (mean $\approx 0.003$, $z=1.28$, $p\approx 0.20$). 

Taken together, conflict energy and NFI indicate that moderate softness ($\tau\approx 1$–$3$) yields controllers that still provide substantial assistance but oppose the user far less often, delivering more “help per unit pushback” than both classic VF and hard-Nash.

\begin{figure}[t]
    \centering
    \includegraphics[width=\linewidth]{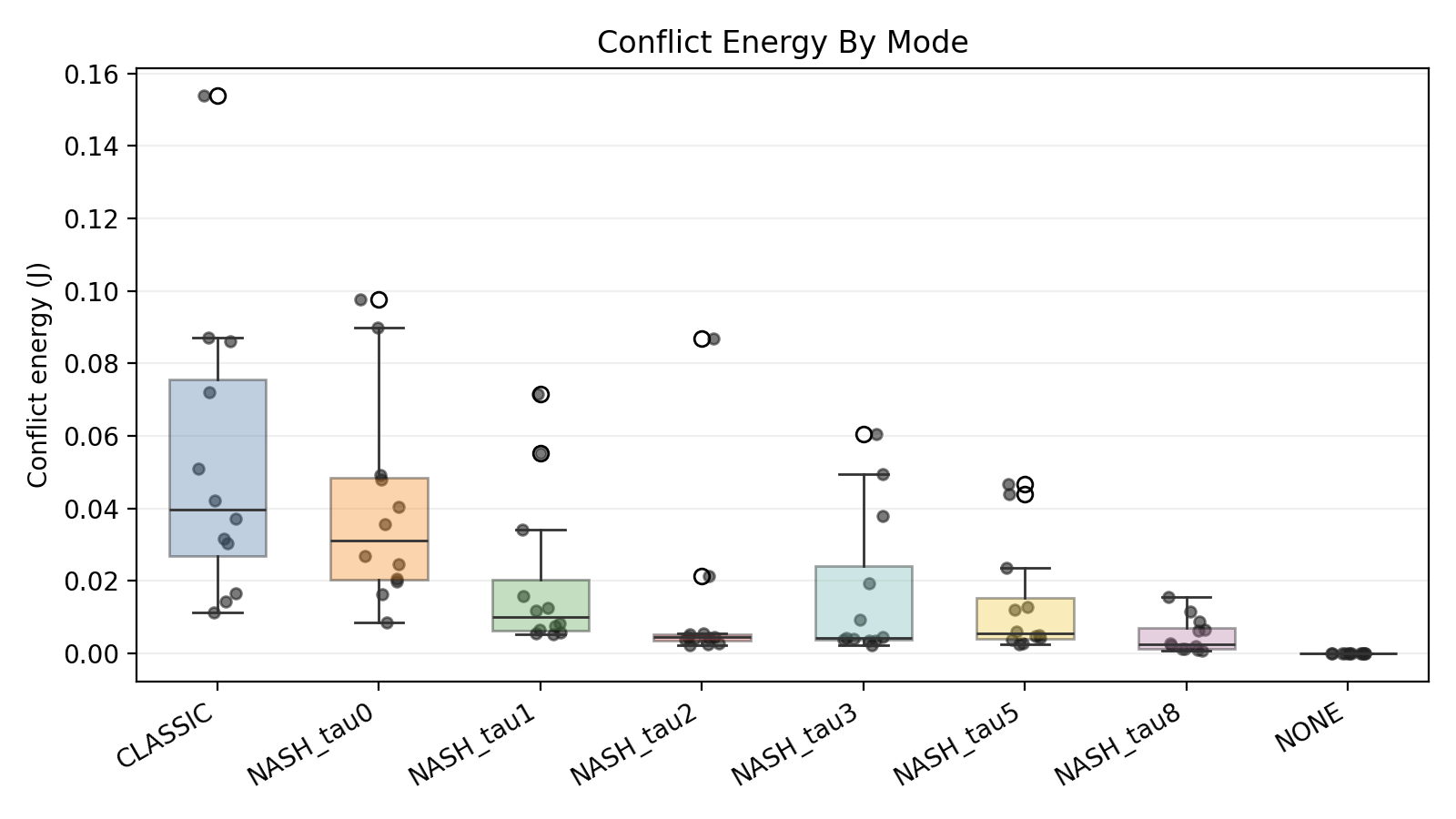}
    \caption{Conflict energy by assistance mode. Classic VF and hard-Nash ($\tau=0$) generate the most controller work directly opposing the user, whereas SoftNash with $\tau \in [1,3]$ reduces conflict by roughly 60–80\% and very soft modes approach the \texttt{NONE} floor.}
    \label{fig:conflict-energy}
\end{figure}

\begin{figure}[t]
    \centering
    \includegraphics[width=\linewidth]{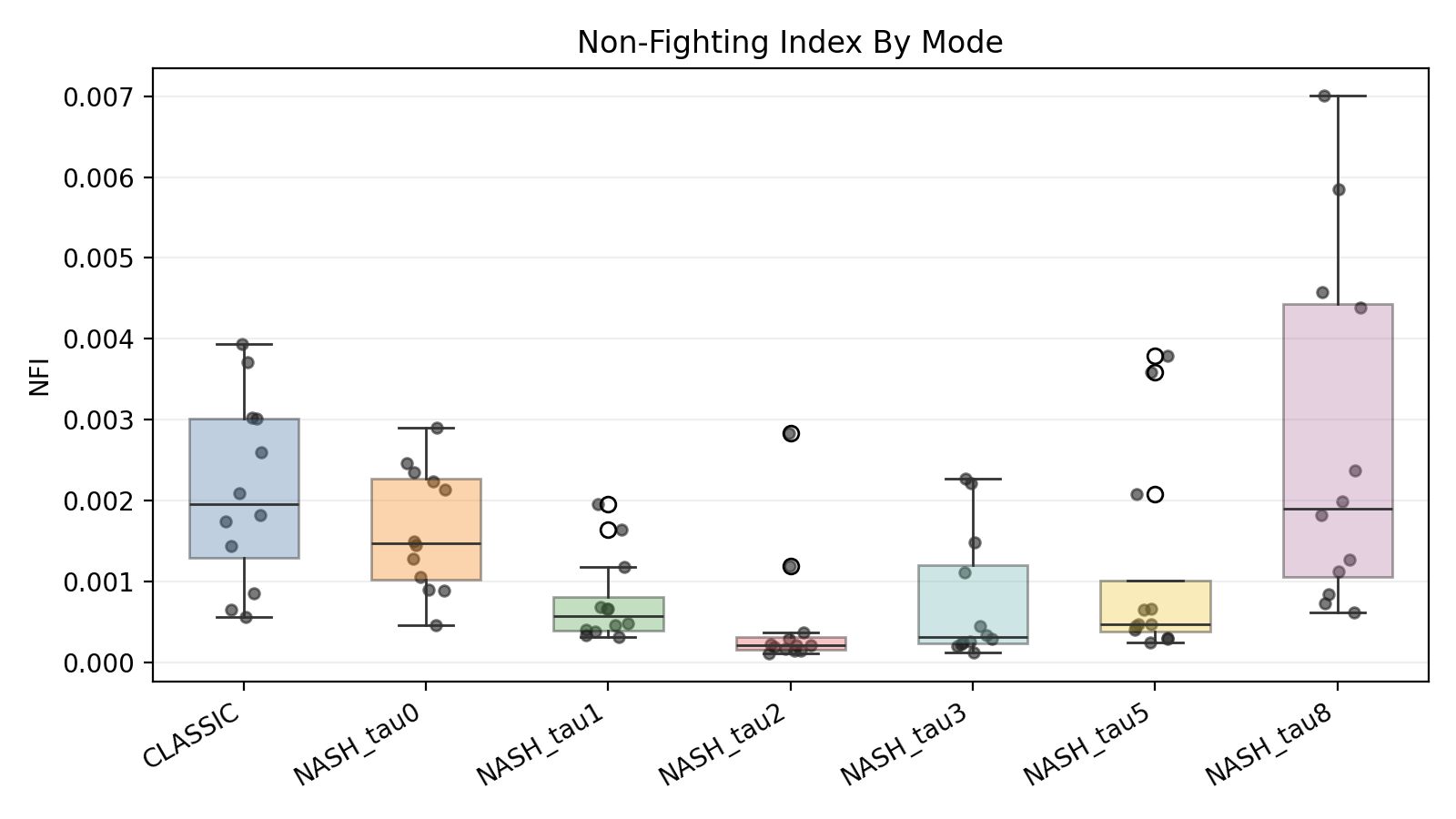}
    \caption{Non-Fighting Index (NFI) by assistance mode. Lower NFI indicates less opposition per unit of delivered assistance; SoftNash with $\tau \in [1,3]$ roughly halves NFI relative to classic VF and hard-Nash, while very soft modes and \texttt{NONE} provide little assistance at all.}
    \label{fig:nfi}
\end{figure}

\subsection{Perceived Agency}

\begin{figure}[t]
    \centering
    \includegraphics[width=\linewidth]{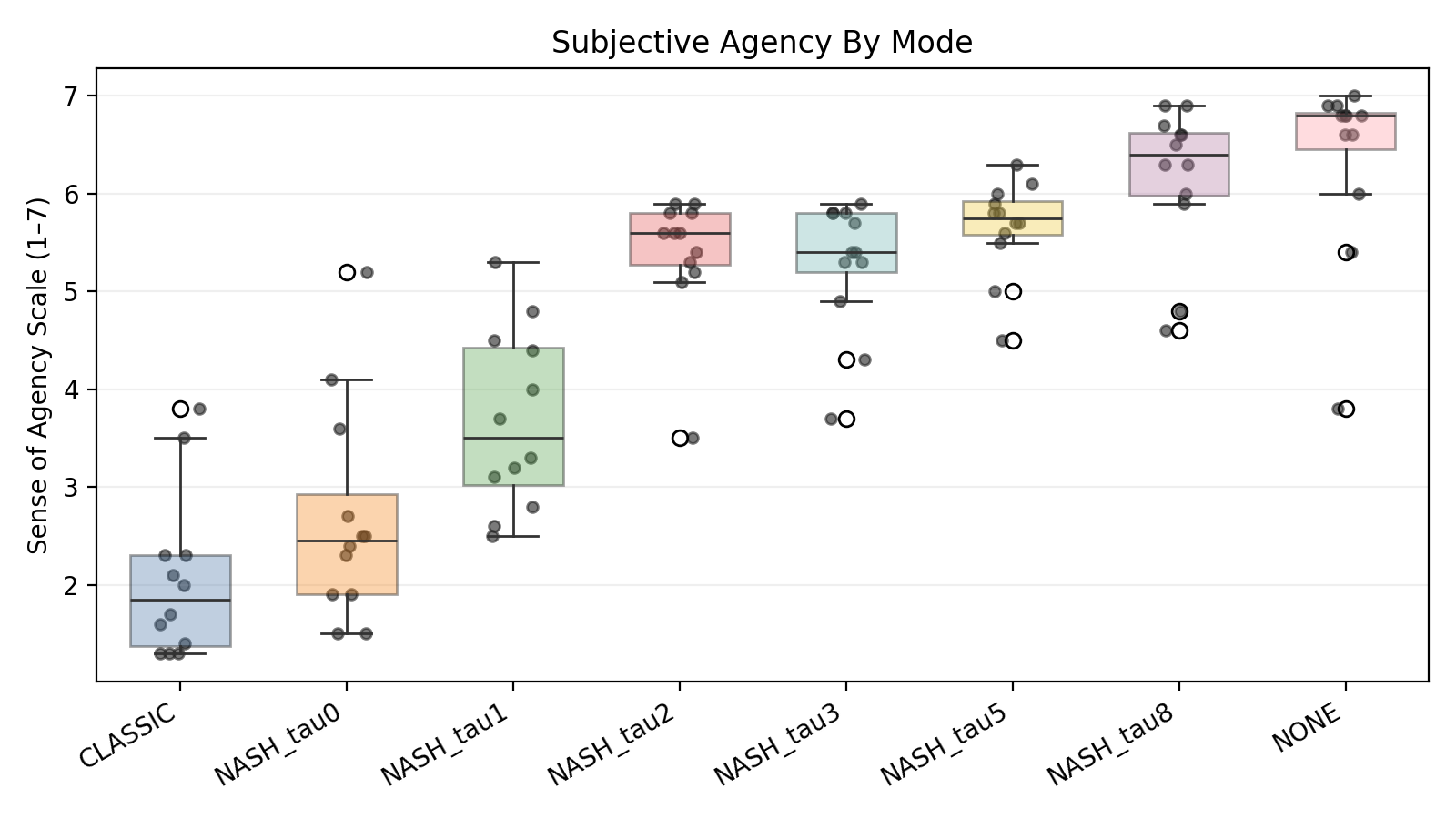}
    \caption{Sense of Agency Scale (SoAS; 1–7) by assistance mode. Perceived agency is lowest for classic VF and hard-Nash, rises sharply for SoftNash around $\tau=2$, and then plateaus, with \texttt{NONE} yielding the highest agency as expected.}
    \label{fig:soas}
\end{figure}

The Sense of Agency Scale (SoAS) exhibited excellent internal consistency (pooled Cronbach’s $\alpha=0.988$, mean per-participant $\alpha=0.982$), indicating reliable measurement across items. A one-way repeated-measures ANOVA revealed a strong effect of mode on SoAS ($F(7,77)=46.81$, $p<0.001$). Classic VF produced low perceived agency (mean $2.05$, 95\% CI $[1.52,\,2.58]$), with hard-Nash slightly higher (mean $2.68$). 

Agency rose steeply with increasing softness: $\tau=1$ yielded mean SoAS $3.68$, and $\tau=2$ jumped to $5.39$ (95\% CI $[4.98,\,5.81]$). Beyond $\tau=2$, improvements were modest: $\tau=3$, $5$, and $8$ produced means of $5.28$, $5.66$, and $6.18$, respectively, with \texttt{NONE} highest at $6.37$. Thus, most of the agency benefit was realized by $\tau\approx 2$, after which the curve essentially plateaued while accuracy began to degrade for very soft controllers.

\subsection{Workload (NASA--TLX)}

\begin{figure}[t]
    \centering
    \includegraphics[width=\linewidth]{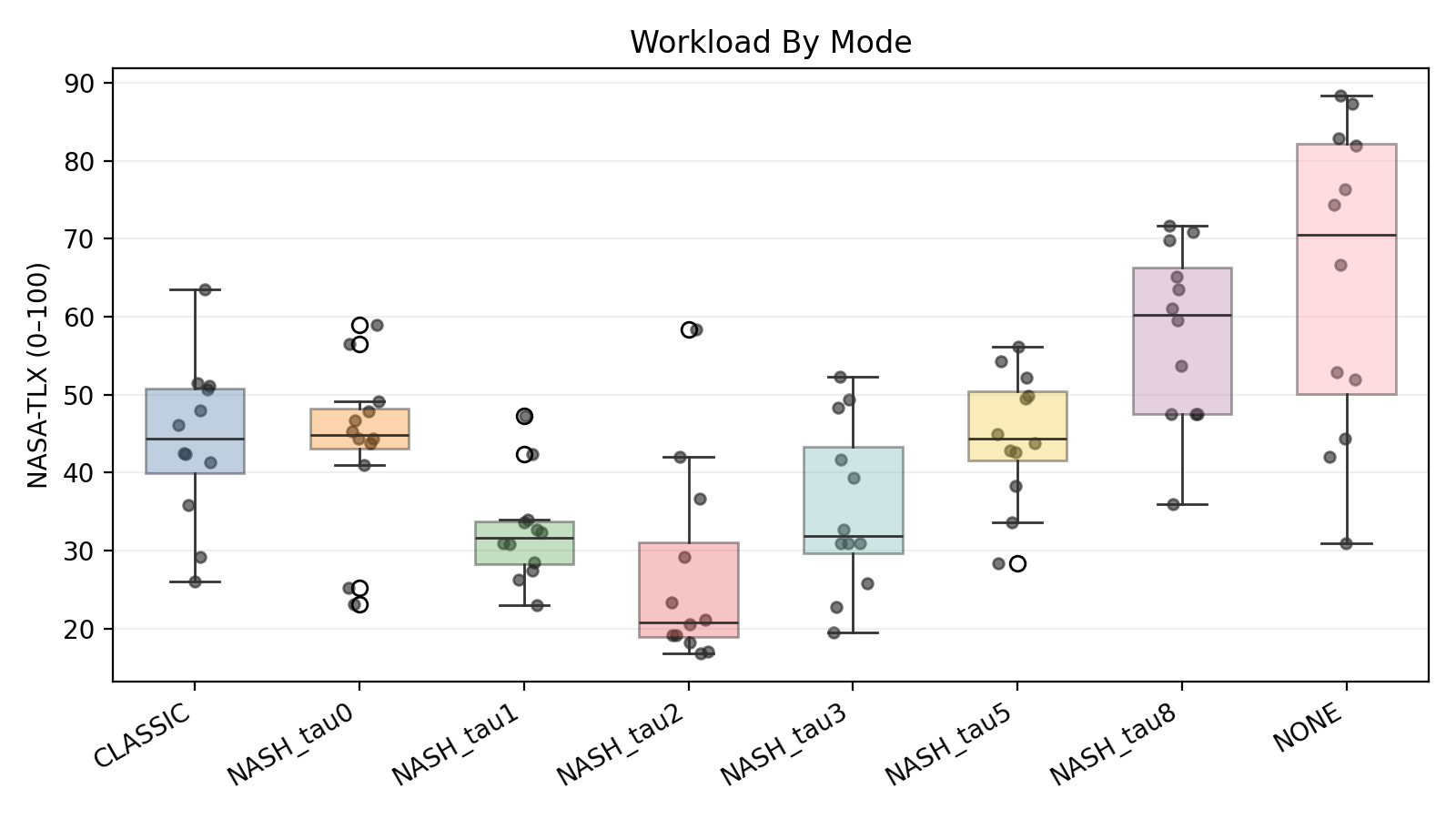}
    \caption{NASA--TLX workload (0–100) by assistance mode. Workload shows a U-shaped dependence on softness: it is highest for \texttt{NONE} and very soft $\tau$ (e.g., 8), moderate for classic VF and hard-Nash, and minimized for SoftNash at $\tau \approx 1$–2.}
    \label{fig:tlx}
\end{figure}

NASA--TLX scores also showed a robust effect of mode ($F(7,77)=19.56$, $p<0.001$). Classic VF and hard-Nash produced similar, moderately high workload (means $44.0$ and $43.9$). SoftNash at $\tau=1$ and $\tau=2$ substantially reduced workload to $32.5$ and $26.8$, respectively (95\% CI for $\tau=2$: $[18.7,\,34.9]$), while still maintaining $\sim 9$~mm RMS error. At $\tau=3$, workload rose again to $35.4$ and by $\tau=5$ ($44.7$) it was comparable to classic VF; $\tau=8$ further increased workload to $57.8$. The \texttt{NONE} condition yielded the highest workload overall (mean $65.0$). 

These trends suggest that moderate SoftNash gains ($\tau\approx 1$–$2$) offload tedious micro-control and reduce cognitive and physical demand without inducing controller fighting, whereas very large $\tau$ or no assistance shift stabilization and planning burdens back to the user.

\subsection{BalancedScore}

\begin{figure}[t]
    \centering
    \includegraphics[width=\linewidth]{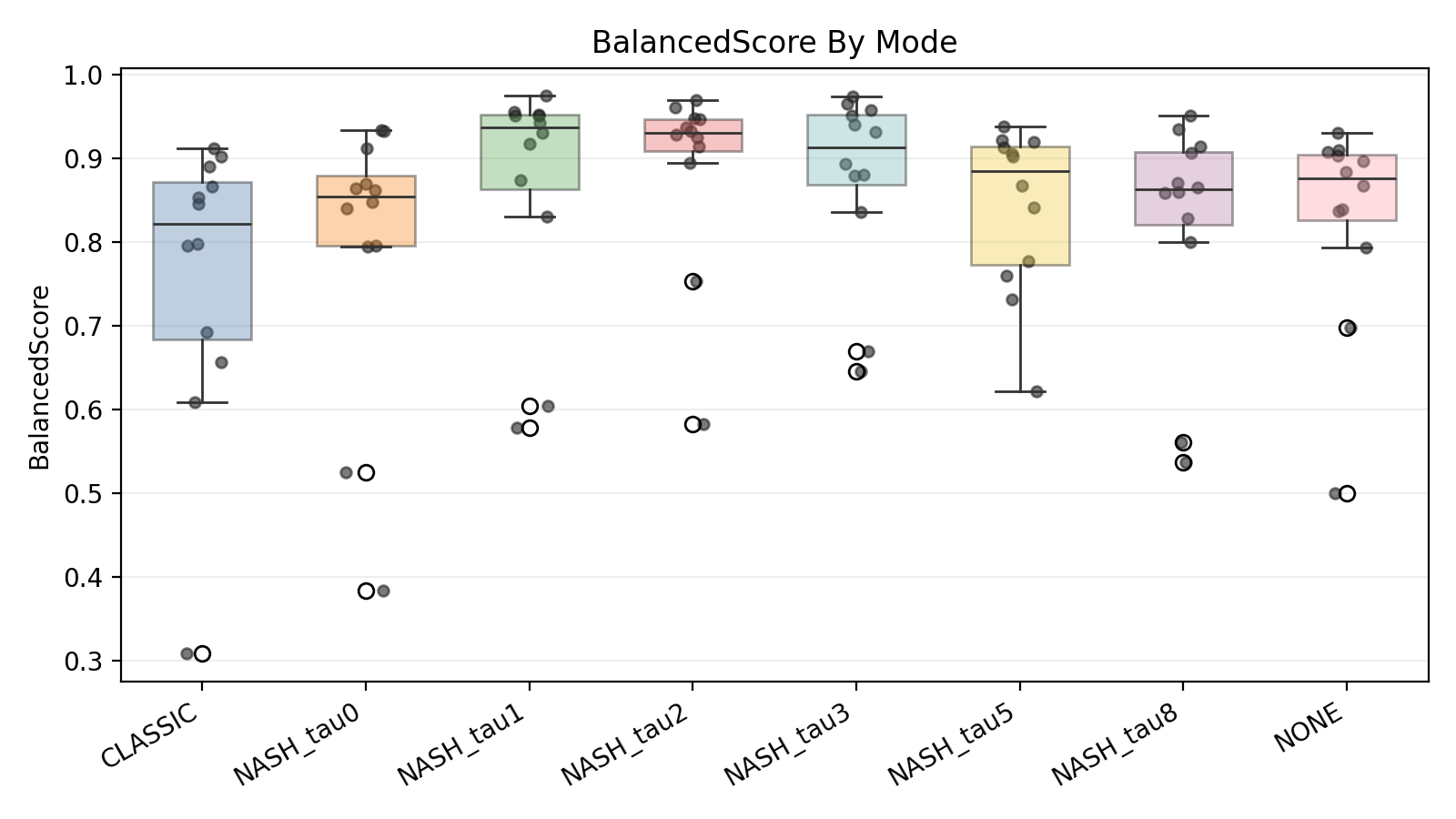}
    \caption{BalancedScore (0–1) by assistance mode, combining normalized tracking accuracy and conflict. The score peaks around SoftNash $\tau=2$–3, indicating a favorable balance between precision and non-fighting behavior, and drops for both over-assertive (classic / $\tau=0$) and under-assistive ($\tau=8$, \texttt{NONE}) regimes.}
    \label{fig:balancedscore}
\end{figure}

BalancedScore, which equally weights normalized tracking accuracy and conflict, was analyzed with a mixed-effects model. Coefficients indicated improvements over classic VF for all SoftNash modes; for example, $\tau=2$ increased BalancedScore by $\beta=0.130$ (SE $\approx 0.026$, $z\approx 4.98$, $p<0.001$). However, planned pairwise contrasts against classic VF were conservative under Holm correction (all adjusted $p>0.18$; nominal $p$ for $\tau\in\{1,2,3\}$ in the range $0.013$–$0.035$). 

Descriptively, BalancedScore peaked near $\tau=2$ (mean $0.891$, 95\% CI $[0.820,\,0.962]$), remained high for $\tau=1$ ($0.872$) and $\tau=3$ ($0.877$), and tapered for larger $\tau$ ($0.842$ at $\tau=5$, $0.824$ at $\tau=8$). Classic VF scored $0.761$, and \texttt{NONE} scored $0.831$. This pattern highlights a favorable operating region around $\tau\approx 2$–$3$ that balances high accuracy with low controller–user conflict.

Overall, the statistical analyses support the qualitative picture from the plots: a moderate SoftNash softness ($\tau \approx 2$) maintains classic VF–level accuracy while significantly improving perceived agency, lowering workload, and reducing conflict per unit of assistance.

\section{Discussion}

Soft-Nash Virtual Fixtures provide a principled and interpretable way to tune shared control. Rather than manually adjusting stiffness or blending gains, designers choose a single parameter $\tau$ that directly trades off between robot aggressiveness and consistency with human input under a Nash game structure.

The empirical results indicate three regimes:
\begin{itemize}
    \item \textbf{Hard regime ($\tau\approx 0$)}: The controller prioritizes performance and behaves like a classic rigid fixture. Accuracy is high, but conflict energy, NFI, workload, and low SoAS scores indicate frequent fighting and reduced agency.
    \item \textbf{Soft regime ($\tau\approx 1$--3)}: The controller maintains near-classic accuracy while substantially reducing conflict and workload, and increasing perceived agency. NFI and BalancedScore suggest that this regime yields the best trade-off between help and non-fighting behavior.
    \item \textbf{Yielding regime ($\tau\gg 1$)}: The controller approaches pass-through behavior, with decreasing conflict but also diminishing performance and increased workload as the human assumes most of the stabilization burden.
\end{itemize}

In this view, $\tau$ acts as a ``gentleness dial'' for the VF: it modulates not only gain magnitude but also the degree of alignment with human intent. The game-theoretic derivation ensures that each setting corresponds to a well-defined equilibrium of an LQ game with regularization, facilitating analysis and extension to richer human models.

\section{Limitations and Future Work}

Our study has several limitations. First, we considered a single device (Omega.6), a single task family (3D point tracking), and a linear--quadratic approximation of the interaction dynamics. Real teleoperation tasks often involve nonlinear dynamics, constraints, and visuohaptic delays that may challenge LQ assumptions.

Second, we represented human intent through a simple affine prior centered on a scaled command and did not adapt $\tau$ online. While this is sufficient to reveal the effect of softness, future work should learn or adapt $\tau$ per user and task, for example via bandit or Bayesian optimization using multi-objective feedback (RMS, NFI, TLX, SoAS).

Third, the experiment was short-term, with a single session per participant. Longitudinal studies are needed to examine whether Soft-Nash VFs preserve or enhance skill transfer and comfort over longer horizons, and whether users evolve interaction strategies under different softness levels.

Finally, while our derivations justify the regularizer in idealized settings, real devices introduce unmodeled dynamics, friction, saturations, and discretization effects. Extending the analysis to robust or stochastic game formulations and validating stability under these non-idealities is an important direction.

\section{Conclusion}

We introduced Soft-Nash Virtual Fixtures, an entropy-regularized two-player LQ game formulation for non-fighting shared control in teleoperation. By inflating the fixture's effort weight with a single parameter $\tau$, we obtain a continuous and interpretable dial on controller assertiveness that preserves Nash equilibrium structure and closed-loop stability continuity. Implementation on a 6-DoF haptic device and evaluation in a 3D tracking task show that moderate softness achieves classic VF-level accuracy while significantly reducing controller--user conflict and workload and increasing perceived agency. These findings support Soft-Nash as a practical, theoretically grounded step toward personalized, agency-preserving shared control in haptics and teleoperation.
\vspace{1cm}

\bibliographystyle{IEEEtran}

\end{document}